\documentclass[letterpaper]{article}
\usepackage{aaai2027}
\usepackage[hyphens]{url}
\usepackage{graphicx}
\usepackage{natbib}
\usepackage{caption}
\usepackage{booktabs}
\usepackage{amsmath}
\usepackage{xcolor}

\title{G-CAST: Governed Continual Self-Training for E-Commerce Agents under Evolving Demands}

\author{
Yuechun Yu\textsuperscript{\rm 1},
Wenzheng Kong\textsuperscript{\rm 1},
Xiaofeng Mao\textsuperscript{\rm 1},
Jin Zhang\textsuperscript{\rm 1}
}
\affiliations{
\textsuperscript{\rm 1}Alibaba Group (\url{alibaba-inc.com})\\
}

\begin{document}
\maketitle
\begin{abstract}



With the rapid advancement of Large Language Models (LLMs), an increasing number of users are turning to LLM-powered agents for conversational shopping in e-commerce. However, the user needs continue to change as new product categories, procurement scenarios, and feature requests emerge. At the same time, the product's execution harness---including its tools, skills, and operating procedures---continues to evolve to support increasingly sophisticated interactions. These changes create two practical challenges: 1) Demand drift alters the distribution, intent, and complexity of user requests, causing a model trained once to fall behind; 2) Harness evolution requires the model to use newly introduced capabilities correctly without losing behavior that users already rely on. Periodic manual retraining is too delayed and labor-intensive for this setting. To address these challenges, we present G-CAST, a governed continual self-training system that converts daily production evidence under evolving demand and capability contracts into reward-routed, entropy-prioritized training data and governed reward evolution. Governed checkpoint transition under cumulative evaluation controls each model update, enabling the agent to learn newly introduced behaviors without sacrificing previously accepted capabilities. Experiments on a real production e-commerce agent show that G-CAST achieves stronger adaptation to newly introduced demand while better preserving previously accepted behavior compared with a strong autonomous post-training baseline. Ablations further show that both reward-routed, entropy-prioritized data orchestration and governed reward evolution are important to these gains. G-CAST is now deployed in Yuancai(http://cai.1688.com/), a B2B procurement agent on Alibaba’s 1688 platform.

\end{abstract}

\section{Introduction}
In e-commerce applications, large language model (LLM)-based shopping agents are deployed to satisfy diverse user requirements and complete a broad range of shopping tasks through conversational interaction. After deployment, user requirements continue to change, becoming increasingly complex and personalized as new product categories, procurement scenarios, and feature requests emerge. To accommodate these changes, the agent harness expands accordingly, incorporating richer context construction, long-term memory, specialized tools, reusable skills, and task-specific operating procedures. These expansions alter the information available to the agent, the actions it can take, and the interaction protocols it is expected to follow. The capability surface exposed to the deployed model therefore changes continually with user requirements.

An expanded capability surface, however, does not automatically lead to improved agent behavior. We refer to the information available to the agent, the actions it can take, and the interaction protocols it must follow as its capability contract. Because the underlying model is typically optimized under an earlier contract, expanding the harness can introduce three coupled mismatches. First, a model mismatch arises when the model fails to recognize when a newly introduced capability is applicable, invoke it correctly, or integrate its output into the final response. Second, a data mismatch arises because trajectories collected under the previous harness may remain useful for legacy capabilities but contain little or no evidence about newly exposed ones. Third, a reward mismatch arises when a reward function designed for the previous contract cannot correctly evaluate behavior under the expanded contract: it may overlook missing or incorrect uses of new tools and skills, or penalize valid behavior that follows a newly introduced procedure.
These mismatches are particularly consequential in e-commerce, where emerging shopping and procurement tasks often require the coordinated use of multiple capabilities. As user requirements and the agent harness evolve, the growing misalignment among the model, training evidence, and reward can prevent the agent from effectively exploiting its expanded capability surface. Harness expansion must therefore be accompanied by continuous agent adaptation. In this setting, self-evolution means continually synchronizing the model, its training evidence, and its reward specification with the evolving capability contract.

Recent approaches to agent self-evolution address individual parts of this problem but leave their coupling unresolved. ACE evolves context playbooks around a fixed model, improving harness-level adaptation without ensuring that the underlying model internalizes newly exposed capabilities \cite{zhang2026ace}. Q-Evolve jointly refines process feedback and the policy, but relies on task-specific environment feedback rather than a capability contract that expands with user requirements \cite{zhang2026qevolve}. SIA combines harness and weight updates, yet still studies improvement within fixed tasks rather than a governed transition that synchronizes learning evidence and reward while protecting previously accepted capabilities \cite{hebbar2026sia}. Consequently, existing methods do not fully address demand-driven self-evolution in deployed agents.

We address this gap with G-CAST, a governed continual self-training system for demand-driven agent adaptation that treats the checkpoint transition, rather than the completion of a training job, as the unit of adaptation. The system jointly versions the deployed model, current capability contract, executable reward, learning evidence, and cumulative evaluation ledger, and couples two mechanisms: reward-routed, entropy-prioritized data orchestration and capability-contract-driven reward evolution. Each round generates multiple rollouts per query with the accepted checkpoint. When their reward patterns provide valid supervision, the data-orchestration mechanism routes all-failure queries with verified targets to supervised fine-tuning, mixed-outcome queries to reinforcement learning, and all-success queries to regression monitoring, then uses token-level entropy to prioritize examples within each training branch. When the reward reliability check instead flags a possible reward error, training is paused and the reward-evolution loop audits the executable reward against the current capability contract. A confirmed mismatch is repaired through an executable reward update that must pass targeted validation, old-versus-new rescoring, and trajectory replay before the evidence is rescored. After these mechanisms establish valid supervision and routed learning evidence, the loop trains a candidate from an immutable parent. The candidate replaces the accepted checkpoint only after passing evaluation on both emerging tasks and all previously retained capabilities. Rejected candidates, regressions, and reward errors guide the next round.

Our contributions are threefold. First, we introduce a governed self-evolution loop that addresses performance degradation caused by increasingly complex user requirements and the corresponding expansion of the agent harness, improving performance on emerging e-commerce tasks while preserving performance on previously accepted tasks. Second, we propose reward-routed, entropy-prioritized data orchestration that decouples the choice of whether and how to train from the prioritization of training examples. By using entropy only to rank uncertainty within routed pools, our method decouples uncertainty from correctness and uses it as a within-pool prioritization signal, resulting in more effective sample selection and better downstream performance compared with alternative strategies. Third, we introduce capability-contract-driven reward evolution with comprehensive gates. As new tools, skills, and operating procedures expand the contract, a fixed reward may retain high scores for trajectories that violate new requirements or penalize behavior that is valid under the expanded contract. Versioned executable rule components keep the reward synchronized, while target-case validation, known-correct regression checks, offline rescoring, and trajectory replay prevent invalid updates from entering training.

\begin{figure*}[ht]
\centering
\includegraphics[width=\textwidth]{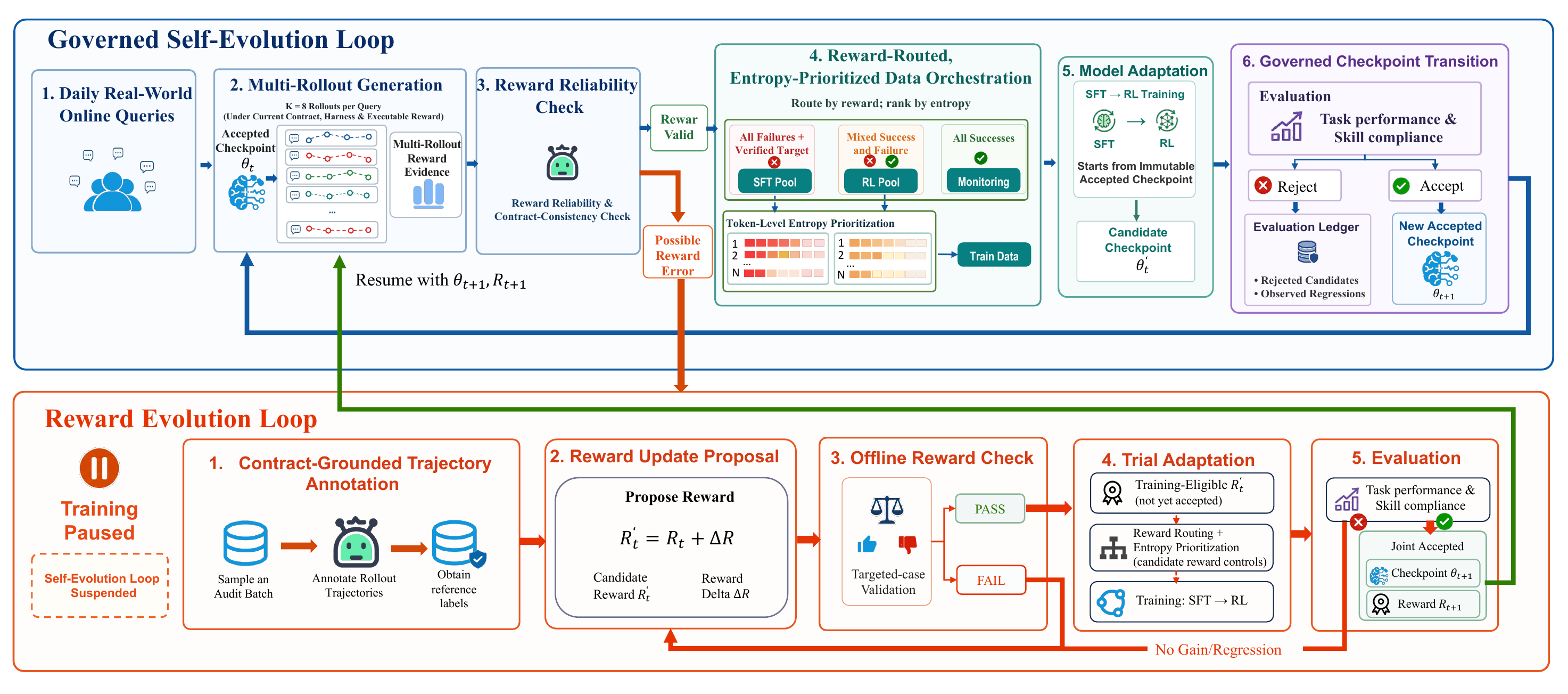}

\caption{G-CAST, a governed continual adaptation loop for deployed e-commerce agents.
The main self-evolution loop converts daily production queries into checkpoint updates through multi-rollout generation, reward reliability checking, reward-routed data orchestration, model adaptation, and governed checkpoint transition. When rollout evidence indicates a possible reward error under the current contract, the reward-evolution loop updates the executable reward through contract-grounded annotation, offline reward checks, and trial adaptation before training resumes. Candidate checkpoints are admitted only after cumulative evaluation on both newly introduced and previously retained capabilities, with rejected candidates, regressions, and reward anomalies recorded in the evaluation ledger.}

\label{fig:loop}
\end{figure*}

\section{Related Work}

Recent work on automated agent improvement follows two main directions. \emph{Automatic training agents} automate post-training and data engineering, whereas \emph{self-evolving agentic AI} updates the context, harness, or model from execution feedback. Most existing methods optimize within a fixed task specification. Our work instead considers evolving user demand, where new capabilities and procedures can simultaneously invalidate model behavior, training evidence, and reward.

\subsection{Automatic Training Agents}

\subsubsection{Autonomous LLM Post-Training}

PostTrainBench and Agent$^2$ RL-Bench evaluate whether LLM agents can autonomously construct post-training pipelines, including data preparation, rollout collection, reward design, training, and evaluation \cite{rank2026posttrainbench,chen2026agent2rlbench}. AutoTrainess improves this process through structured interfaces and operational constraints, while A-Evolve-Training demonstrates autonomous post-training at larger model scales \cite{yu2026autotrainess,shi2026aevolve}. These systems optimize a predefined task objective; our setting additionally requires determining whether the objective and reward remain valid as user requirements evolve.

\subsubsection{Autonomous Data Engineering}

ADP-MA automates multi-stage data processing through hierarchical meta-agents, schema contracts, and runtime validation \cite{khurana2026adpma}. DataMaster searches over data discovery, selection, and transformation using shared data and cumulative memory \cite{du2026datamaster}. CURATION-BENCH studies agent-generated data-selection policies, while Autodata constructs and iteratively refines synthetic datasets \cite{kang2026curationbench,kulikov2026autodata}. Unlike these methods, which improve data under a fixed learning objective, our method validates the current reward and routes production evidence to SFT, RL, or regression monitoring according to repeated-rollout outcomes.

\subsection{Self-Evolving Agentic AI}

\subsubsection{Harness and Context Evolution}

Harness-level approaches adapt external components while keeping model parameters fixed. ACE evolves context playbooks from execution feedback \cite{zhang2026ace}; Meta Context Engineering co-evolves context-engineering skills and context artifacts \cite{ye2026mce}; and Meta-Harness searches over executable agent scaffolds \cite{lee2026metaharness}. However, an improved harness does not guarantee that the underlying model can exploit it. Recent work distinguishes the ability to produce useful harness updates from the ability to benefit from them \cite{lin2026harnessbenefit}, motivating weight adaptation when newly introduced capabilities exceed the behavior learned under the previous harness.

\subsubsection{Joint Harness and Model Evolution}

SIA jointly updates a task-specific harness and model weights from verifier feedback, making it the closest prior work in update scope \cite{hebbar2026sia}. However, it assumes a given task specification and verifier. Our work begins when evolving demand expands the capability contract itself, potentially creating coupled model, data, and reward mismatches. We therefore treat the checkpoint transition as the unit of self-evolution: the loop validates contract-aware rewards, routes rollout evidence to SFT or RL, trains from an immutable accepted parent, and promotes a candidate only after cumulative evaluation on both emerging and previously accepted tasks.

\section{Application}

Our study is applied in Yuancai (\url{http://cai.1688.com/}), a professional AI agent product for business-to-business (B2B) buyers built on 1688, Alibaba's wholesale sourcing platform. Yuancai pairs LLM-based agents with 1688's factory and supply-chain resources, and its purchasing capabilities have evolved over time as new operating modes, tools, and procedures were introduced in response to changing business needs. A general assistant mode handles arbitrary multi-step sourcing and business requests described in a single instruction, planning and executing over more than twenty built-in tools with real-time scheduling, multi-turn reasoning, and automatic self-correction. A quick search mode serves precise product, supplier, and look-alike retrieval through a shorter execution path with faster response. A knowledge-assisted mode conditions execution on configured business playbooks, uploaded documents, and accumulated interaction history, continuously memorizing know-how from ongoing conversations while retaining and upgrading prior buyer-workspace functions such as instant messaging and web browsing. Each addition expanded the context, tools, procedures, and outputs available to the model, thereby creating a new operating contract at that stage of evolution.

These evolving functions provide the real-world setting for G-CAST. Each newly introduced capability defines a revised operating contract together with a corresponding slice of production demand. This gives us a sequence of versioned demand-and-contract snapshots on which continual adaptation can be studied and governed.

\begin{table}[h]
\centering
\small
\begin{tabular}{lrl}
\toprule
Period & Tools & Newly required capability \\
\midrule
Period 1 & 4 & basic retrieval and detail lookup \\
Period 2 & 7 & external evidence and trend lookup \\
Period 3 & 17 & skills, multimodal, inquiry workflows \\
Period 4 & 23 & sandbox, files, export, full workflows \\
\bottomrule
\end{tabular}
\caption{Demand contexts sampled from four distinct time periods. Each period defines a frozen snapshot of requests and capability requirements observed during its corresponding time span, rather than a stage of the learning algorithm.}
\label{tab:contexts}
\end{table}

\section{Method}

Figure~\ref{fig:loop} illustrates G-CAST as a governed continual self-training system for deployed agents whose user demand and capability contract evolve over time. In this setting, adaptation is not just another training run: newly introduced tools, skills, or procedures can simultaneously create model mismatch, data mismatch, and reward mismatch. G-CAST addresses these coupled failures with two linked processes in Figure~\ref{fig:loop}: a main self-evolution loop that turns validated production evidence into checkpoint updates, and a reward-evolution loop that repairs reward when contract change makes the current reward unreliable.

In the main loop, each round starts from one immutable accepted checkpoint under the current harness and executable reward. Selected production queries are replayed to generate multiple rollouts, which then pass through a reward reliability check that separates training-eligible evidence from regression-monitoring evidence and flags possible reward errors. Training-eligible evidence is then processed by reward-routed, entropy-prioritized data orchestration, while possible reward errors trigger the reward-evolution loop and, if needed, executable reward update before training proceeds. The loop then trains a candidate checkpoint from the accepted parent and admits the checkpoint transition only if cumulative evaluation confirms both adaptation to newly introduced demand and retention of previously accepted behavior. Table~\ref{tab:contexts} summarizes the period-indexed capability contracts used throughout this process.

\paragraph{Reward-Routed, Entropy-Prioritized Data Orchestration.}

The reward-routed, entropy-prioritized data orchestration component in Figure~\ref{fig:loop} determines how each query contributes to self-training. For each query, the accepted checkpoint generates eight rollouts. Queries whose rollouts all fall below a round-level reference enter the SFT pool when a verified target exists. Queries with both above- and below-reference rollouts enter the RL pool, since they provide outcome variation suitable for relative optimization. Queries whose rollouts all remain above the reference are retained for regression monitoring. The reference is defined as the mean valid reward after trimming the top and bottom 5\%.

After routing, G-CAST prioritizes examples within the SFT and RL pools by token-level uncertainty. For each rollout, we compute the 90th percentile of token entropies and define the query score as the maximum across its rollouts. Reward decides the learning role of each query, while entropy focuses the training budget on the most informative routed examples.

\paragraph{Contract-Driven Reward Evolution.}

The reward-evolution loop in Figure~\ref{fig:loop} is triggered when the reward reliability check indicates that the current executable reward is not scoring trajectories reliably for the current deployment setting. This can happen when the capability contract expands to new tools, skills, or procedures, or when audit evidence reveals a scoring mismatch under the existing contract, such as high reward on trajectories that should be penalized. G-CAST therefore treats reward as a governed executable object that must remain aligned with both the active capability contract and the distinction between desirable and undesirable trajectories.

The loop follows the five stages in Figure~\ref{fig:loop}. Contract-grounded trajectory annotation collects audit cases and reference labels for the suspected failure mode. Reward update proposal then specifies the executable rule change needed to correct that mismatch. Offline reward validation tests the proposed reward through targeted cases, regression checks, offline rescoring, and replay or dry-run execution, verifying that it improves the scoring of desirable versus undesirable trajectories without breaking previously correct judgments. Trial adaptation uses the candidate reward to control routing and training on an audit batch. Finally, joint evaluation determines whether the updated reward and the resulting candidate checkpoint should be admitted; otherwise, the loop resumes with the previously accepted reward and checkpoint.

Within this loop, the executable reward is a versioned bundle of components. Each component has an explicit applicability condition. A proposed reward update must specify the scoring error it corrects and, when relevant, the contract change it encodes, and must pass validation before being used for training.

Following the capability expansion in Table~\ref{tab:contexts}, the training
reward evolves cumulatively with the period-specific contract:
\begin{align}
R_{\mathrm{train}}^{(1)}
&=R_{\mathrm{fmt}},\\
R_{\mathrm{train}}^{(2)}
&=R_{\mathrm{fmt}}+P_{\mathrm{gd}}+P_{\mathrm{retrieval}},\\
R_{\mathrm{train}}^{(3)}
&=R_{\mathrm{fmt}}+P_{\mathrm{gd}}+P_{\mathrm{retrieval}}
+R_{\mathrm{skill\mbox{-}sel}}+R_{\mathrm{skill\mbox{-}adh}},\\
R_{\mathrm{train}}^{(4)}
&=R_{\mathrm{train}}^{(3)} 
=R_{\mathrm{full}}
\end{align}
Thus, from Period~3 onward, the training reward is the full reward
$R_{\mathrm{full}}$. Period~4 does not introduce a new scalar reward term; instead, it retains the same full reward while expanding the applicability and executable checks of the
existing components to sandbox, file-processing, export, and longer workflows.
Each term contributes only when it is applicable to the query contract.
$R_{\mathrm{fmt}}$ measures rollout-level compliance with the required
output and interaction rules rather than task correctness. It starts from a positive base score and is reduced by
malformed or unparseable tool actions, redundant repeated tool calls, answers
returned without required business-tool use, invalid actions after termination,
or missing final answers. $R_{\mathrm{skill\mbox{-}sel}}$ evaluates whether
the required skill is correctly selected against a teacher-verified reference,
rewarding correct selection and lowering the score for missing, incorrect, or
unnecessary skill calls. $R_{\mathrm{skill\mbox{-}adh}}$ is rule-based and
checks whether the rollout follows the procedure specified in the selected
\texttt{SKILL.md}, including the required tool sequence when applicable.
$P_{\mathrm{gd}} \leq 0$ captures grounding quality by lowering the score when
factual content in the final answer is not supported by the user prompt or
returned tool evidence. $P_{\mathrm{retrieval}} \leq 0$ applies when a final
answer claims retrieval failure without providing any valid item or shop
result.

The two skill-related terms
are introduced when later contracts require explicit skill use and procedure
compliance (Period 3-4). The grounding penalty is another evolved component, added after
reward audits identified unsupported answer content. Before being used for
training, each reward update must pass offline reward check. For analyses that exclude the two
skill-related terms, we additionally report
\begin{equation}
R_{\mathrm{base}}
=R_{\mathrm{full}}
-R_{\mathrm{skill\mbox{-}sel}}
-R_{\mathrm{skill\mbox{-}adh}}.
\end{equation}
This base reward is an analytical decomposition used for comparison across
contract stages, not an earlier historical reward version.

\paragraph{Evaluation metrics.}
Unless explicitly noted as a base-reward analysis, all reported reward metrics
use the mean frozen full evaluation reward. Let $E_j$ denote the held-out
evaluation slice for period $j$, let $R_{\mathrm{full}}$ denote the full reward of query
$q$, and let $t$ denote the current stage. We define
\begin{align}
R_{\mathrm{ret}}^{(t)}
&=\frac{\sum_{j<t}\sum_{q\in E_j}R_{\mathrm{full}}}
        {\sum_{j<t}|E_j|}, \qquad t>1, \\
R_{\mathrm{new}}^{(t)}
&=\frac{1}{|E_t|}\sum_{q\in E_t}R_{\mathrm{full}}, \\
R_{\mathrm{cum}}^{(t)}
&=\frac{\sum_{j\leq t}\sum_{q\in E_j}R_{\mathrm{full}}}
        {\sum_{j\leq t}|E_j|}.
\end{align}
Retained-period reward $R_{\mathrm{ret}}^{(t)}$ measures performance on previously introduced demand
periods, new-period reward $R_{\mathrm{new}}^{(t)}$ measures performance on the newly introduced period,
and cumulative reward $R_{\mathrm{cum}}^{(t)}$ measures overall performance on all active periods.

\paragraph{Frozen evaluation reward.}
The training reward evolves with the capability contract during the loop, but all results reported in this paper are measured with one common frozen evaluation reward: the final contract-complete version. It combines always-applicable task-outcome and format components, skill-compliance components that apply only when the query contract requires a skill, and the gate-admitted grounding penalty. Applicability is determined solely by the query contract, never by the checkpoint or ablation arm being scored. This common frozen measure keeps all reported comparisons directly comparable and evaluates every checkpoint under the same deployment-relevant contract.

\section{Experiments}
\label{sec:experiments}

 This section evaluates G-CAST on a production e-commerce agent under evolving demand and capability contracts. Section 5.1 describes the experimental setup, including the four demand periods, evaluation protocol, and common frozen evaluation reward. Section 5.2 studies adaptation and retention, comparing G-CAST with AutoTrainess under matched cumulative evaluation. Section 5.3 presents the routing and prioritization ablation, isolating the effects of reward routing and entropy-based within-pool ranking. Section 5.4 evaluates reward evolution and examines whether governed reward updates are necessary under an expanding capability contract. Section 5.5 reports operational impact beyond evaluation reward, including maintenance efficiency and governance outcomes.

\subsection{Experimental Setup}

We evaluate G-CAST on representative production queries sampled from four distinct time periods of the deployed application. These periods define four disjoint demand contexts that reflect evolving product categories, procurement scenarios, interaction patterns, and capability requirements. Each context contains a frozen 2{,}000-query rollout pool and a disjoint 100-query evaluation slice created before adaptation. AutoTrainess is evaluated on the same four frozen contexts and the same cumulative evaluation slices, but follows its own stage-wise training policy for data selection.

We use Qwen3-4B-Thinking-2507 as the pretrained model. For each query in a rollout pool, the accepted checkpoint generates 8 trajectories under the corresponding harness version, yielding 16{,}000 trajectories per context before routing. When a new context is introduced, the accepted checkpoint is first evaluated on that period's 100-query slice. Every subsequent candidate is then evaluated on the cumulative suite of all active slices, which grows from 100 to 400 queries across the four periods. This cumulative protocol is used for both G-CAST and AutoTrainess and matches the deployment requirement: a candidate must address newly emerging demand without sacrificing previously accepted capabilities.

To ensure fair comparison, unless explicitly varied by an ablation, all methods within an experiment use the same parent model, harness version, candidate universe, teacher targets, and decoding configuration. Interaction records are deduplicated and normalized while preserving their source context, tool evidence, and harness provenance. We also perform data decontamination to prevent evaluation leakage: evaluation queries are excluded from all training, replay, synthesis, and reward-development processes, ensuring strict separation between training and test data.


\begin{figure}[t]
\centering
\includegraphics[width=\columnwidth]{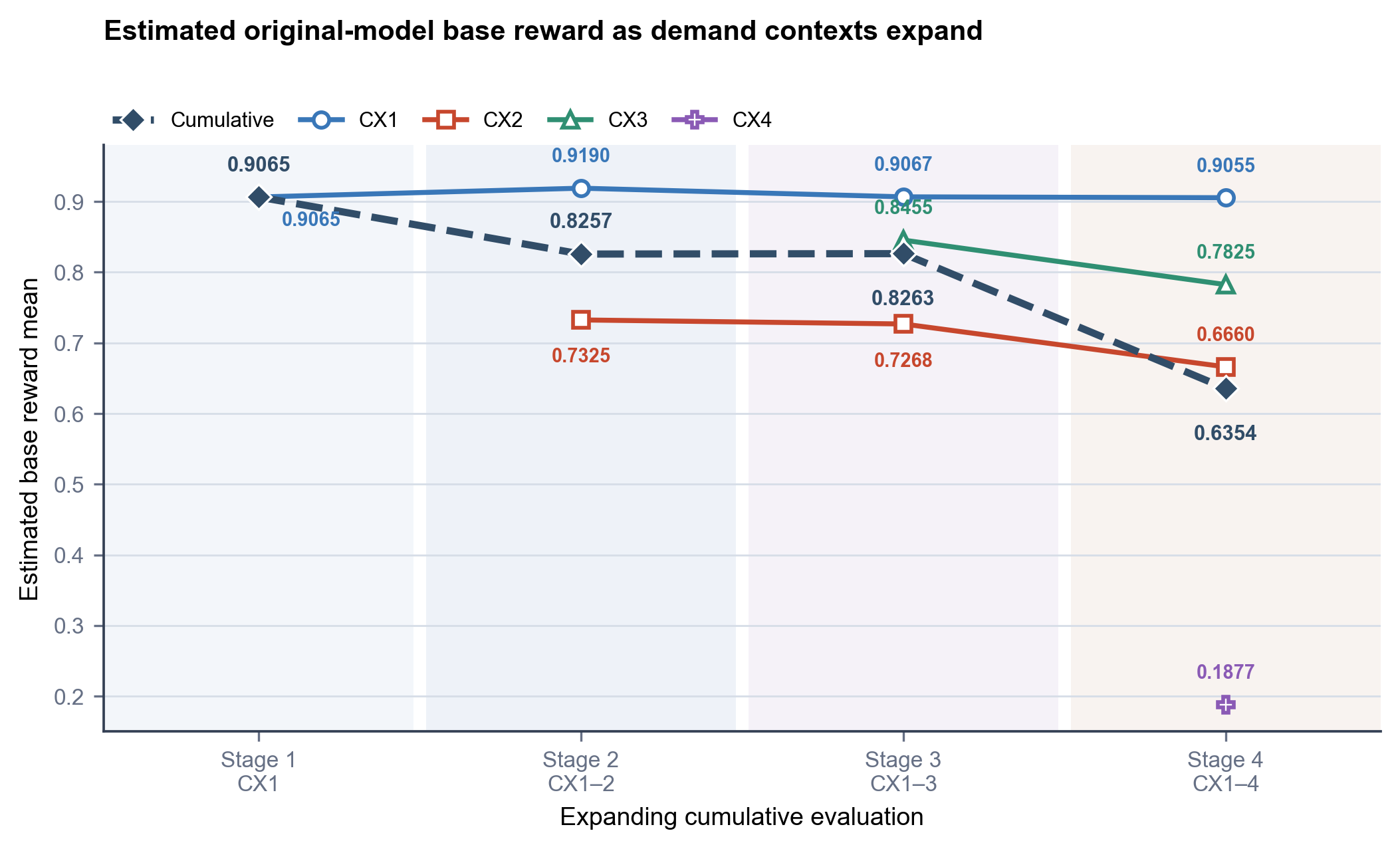}
\caption{Performance degradation of a fixed checkpoint as the capability contract expands across demand periods.
The figure reports the base reward, which excludes auxiliary reward components introduced only in later contract versions, so that performance remains directly comparable across periods. The dark curve shows the cumulative mean over all active evaluation slices, and the colored curves show the period-specific means.}
\label{fig:cleanref-degradation}
\end{figure}

\begin{table*}[t]
\centering
\scriptsize
\setlength{\tabcolsep}{2.1pt}
\renewcommand{\arraystretch}{1.12}
\resizebox{\textwidth}{!}{%
\begin{tabular}{cclccccccc}
\toprule
Iter. & Evaluation & Method
& SFT / RL rows
& \multicolumn{4}{c}{Cumulative reward}
& \multicolumn{2}{c}{From stage entry to Post-RL} \\
\cmidrule(lr){5-8}\cmidrule(lr){9-10}
& & &
& Base & Entry  & Post-SFT & Post-RL
& \shortstack{Retained periods\\before $\rightarrow$ after ($\Delta$)}
& \shortstack{New period\\before $\rightarrow$ after ($\Delta$)} \\
\midrule
1 & P1 (100 queries) & AutoTrainess & -- / 1200
& 0.9065 & 0.9065 & -- & \textbf{0.9360}
& -- & 0.9065\thinspace$\rightarrow$\thinspace0.9360\thinspace(\textbf{+0.030}) \\
  &                  & G-CAST       & 94 / 500
& 0.9065 & 0.9065 & 0.9180 & 0.9240
& -- & 0.9065\thinspace$\rightarrow$\thinspace0.9240\thinspace(+0.018) \\
\midrule
2 & P1--2 (200 queries) & AutoTrainess & -- / 1200
& 0.8258 & 0.8665 & -- & \textbf{0.8935}
& 0.9365\thinspace$\rightarrow$\thinspace0.9325\thinspace(\textcolor{red}{$-$0.004})
& 0.7965\thinspace$\rightarrow$\thinspace0.8545\thinspace(+0.058) \\
  &                     & G-CAST       & 54 / 984
& 0.8258 & 0.8253 & 0.8473 & 0.8835
& 0.9240\thinspace$\rightarrow$\thinspace0.9300\thinspace(\textbf{+0.006})
& 0.7265\thinspace$\rightarrow$\thinspace0.8370\thinspace(\textbf{+0.110}) \\
\midrule
3 & P1--3 (300 queries) & AutoTrainess & -- / 1200
& 0.8568 & 0.8876 & -- & 0.9210
& 0.8700\thinspace$\rightarrow$\thinspace0.9119\thinspace(+0.042)
& 0.9227\thinspace$\rightarrow$\thinspace0.9392\thinspace(+0.017) \\
  &                     & G-CAST       & 62 / 778
& 0.8568 & 0.8453 & 0.8581 & \textbf{0.9288}
& 0.8198\thinspace$\rightarrow$\thinspace0.9163\thinspace(\textbf{+0.097})
& 0.8964\thinspace$\rightarrow$\thinspace0.9538\thinspace(\textbf{+0.057}) \\
\midrule
4 & P1--4 (400 queries) & AutoTrainess & -- / 247
& 0.6501 & 0.7538 & -- & 0.7911
& 0.8907\thinspace$\rightarrow$\thinspace0.8739\thinspace(\textcolor{red}{$-$0.017})
& 0.3433\thinspace$\rightarrow$\thinspace0.5428\thinspace(+0.200) \\
  &                     & G-CAST       & 122 / 859
& 0.6501 & 0.7530 & -- & \textbf{0.8078}
& 0.8712\thinspace$\rightarrow$\thinspace0.8833\thinspace(\textbf{+0.012})
& 0.3983\thinspace$\rightarrow$\thinspace0.5813\thinspace(\textbf{+0.183}) \\
\bottomrule
\end{tabular}%
}
\caption{Cumulative performance, retention, and adaptation under matched
evaluation. The four middle columns report cumulative reward at successive
checkpoints. In each of the two rightmost columns, a cell reports the
method-specific stage-entry value, the Post-RL value, and their difference.
Retained reward is the
query-weighted mean over previously introduced 100-query periods, whereas
new-period reward is computed on the newly introduced 100-query period. A
negative retained delta denotes forgetting. "–" indicates a non-applicable metric or an unreported intermediate result.}
\label{tab:main-loop-results}
\end{table*}

\subsection{Continual Self-Training under Evolving Contracts}

We first verify that continual self-training is necessary in this setting. Figure~\ref{fig:cleanref-degradation} evaluates the original checkpoint without updating it as the application evolves from Period 1 to Period 4. To keep the degradation trend directly comparable across evolving contracts, we report the base reward in the main text. Under this metric, the checkpoint’s cumulative performance degrades substantially as the contract expands, falling from 0.9065 to 0.6354 on the cumulative Periods 1–4 evaluation, while its reward on the newly introduced Period 4 falls to 0.1877. This result shows that a checkpoint optimized under an earlier contract does not automatically benefit from later harness expansion. Detailed period-wise values and full metric breakdowns are deferred to Appendix Table 2.

Against this background, Table~\ref{tab:main-loop-results} compares G-CAST with AutoTrainess under the same cumulative evaluation protocol. In the first two iterations, G-CAST already achieves strong gains with substantially less training data. In Iteration~1, G-CAST uses 94 SFT queries and 500 RL queries, compared with 1200 training queries for AutoTrainess, and improves cumulative reward from 0.9065 to 0.9240. In Iteration~2, G-CAST again trains on fewer queries (54 SFT + 984 RL versus 1200) and reaches 0.8835. These early results show that governed routing can convert a relatively small amount of production evidence into effective continual self-training.

The more important difference emerges once the capability contract expands to require explicit skill-dependent behavior. In Iteration~3, G-CAST improves cumulative reward from 0.8453 at stage entry to 0.9288 after RL, surpassing AutoTrainess at 0.9210. In Iteration~4, under the fullest contract expansion, G-CAST again achieves the stronger cumulative result, reaching 0.8078 versus 0.7911 for AutoTrainess. This crossover is the central result of the continual self-training study: when the loop must track not only new demand but also a changing capability contract, governed reward and data evolution become more important than broad post-training under a fixed objective alone.

The last two columns of Table~\ref{tab:main-loop-results} further show that G-CAST delivers a stronger adaptation--retention profile. In Iteration~4, for example, G-CAST improves new-period reward from 0.3983 to 0.5813 while also increasing retained reward from 0.8712 to 0.8833. AutoTrainess also improves the new period, from 0.3433 to 0.5428, but its retained reward decreases from 0.8907 to 0.8739. The advantage of G-CAST is therefore not only higher cumulative reward, but a more favorable checkpoint transition under cumulative deployment constraints.

This difference is concentrated in the newly important contract behaviors. In Iteration~3, G-CAST reaches higher skill-selection accuracy (0.8667 vs.\ 0.8433) and higher skill adherence (0.5233 vs.\ 0.5153). In Iteration~4, where correct skill usage becomes more consequential, the margin widens further: G-CAST attains 0.7625 skill-selection accuracy versus 0.6875 for AutoTrainess, and 0.4503 skill adherence versus 0.4219. These results are consistent with the intended role of G-CAST: not to optimize for a single static objective, but to keep the deployed agent aligned with newly introduced capabilities while preserving behavior that earlier users already rely on.

 The smaller AutoTrainess training set in Iteration 4 reflects its own targeted stage objective rather than reduced access to the rollout pool (Appendix B). Appendix Tables 2 and 3 provide the full retained reward, new-period reward, cumulative reward, format reward, skill-selection accuracy, and skill adherence breakdowns across the continual self-training process.




\subsection{Routing and Prioritization Ablation}

\begin{table*}[ht]
\centering
\small
\setlength{\tabcolsep}{8pt}
\begin{tabular}{llrrrr}
\toprule
& & \multicolumn{2}{c}{Task Performance} & \multicolumn{2}{c}{Skill Compliance} \\
\cmidrule(lr){3-4}\cmidrule(lr){5-6}
Stage & Data strategy & Cumulative reward $\uparrow$ & Format $\uparrow$ & Sel. Acc. $\uparrow$ & Adherence R. $\uparrow$ \\
\midrule
Base & Qwen3-4B-Thinking-2507-FP8 & 0.7331 & 0.7368 & 71.50\% & 0.3446 \\
\midrule
Post-SFT & Random & 0.7200 & 0.7300 & 71.50\% & 0.3567 \\
Post-SFT & Reward + Random & 0.7361 & 0.7466 & 72.25\% & 0.3570 \\
Post-SFT & Reward + Entropy (Ours) & 0.7800& 0.7831 & 73.25\% & 0.3534 \\
\midrule
Post-RL & Random & 0.7334 & 0.7520 & 70.00\% & 0.3504 \\
Post-RL & Reward + Random & 0.7721 & 0.7814 & 73.00\% & 0.3541 \\
Post-RL & Reward + Entropy (Ours) & \textbf{0.7930} & \textbf{0.8069} & \textbf{73.25\%} & \textbf{0.3637} \\
\bottomrule
\end{tabular}
\caption{Routing and prioritization ablation on the cumulative evaluation across Periods~1--4. Reward + Entropy is the full method, Reward + Random removes entropy prioritization, and Random removes both components. Sel. Acc. denotes skill-selection accuracy, and Adherence R. denotes skill-adherence reward. }
\label{tab:routing-results}
\end{table*}

While Table~\ref{tab:main-loop-results} establishes the overall advantage of G-CAST under cumulative deployment constraints, it does not by itself explain which part of the loop is responsible for that gain. In particular, G-CAST differs from broad post-training baselines in how production evidence is converted into training signal. The next experiment therefore isolates the data-orchestration component under the same candidate universe and training budget, evaluating whether reward-based routing improves sample utility and whether entropy-based ranking provides additional benefit after routing has determined the learning role of each query.

We evaluate the data-orchestration component with three matched selection arms. \textbf{Random} applies no reward routing and samples directly from the common candidate universe. \textbf{Reward + Random} applies the reward-routing rule from the Method section and samples randomly within the resulting SFT and RL pools. \textbf{Reward + Entropy} applies the same routing rule and ranks examples within each pool by the entropy-based uncertainty score. Random versus Reward + Random isolates the effect of reward routing, while Reward + Random versus Reward + Entropy isolates the additional effect of entropy-based prioritization.

Each arm selects 150 SFT queries and 500 RL queries. It first performs SFT and then continues RL from that arm's own SFT checkpoint. The arms match the base model, candidate universe, data budget, SFT targets, training configuration, and evaluation protocol; only routing and within-pool selection differ. We report both post-SFT and post-RL checkpoints, and use a length-matched analysis to test whether entropy acts merely as a proxy for response length.

Table~\ref{tab:routing-results} reports the post-SFT and post-RL results under three data strategies. Reward mean reflects overall task performance by averaging trajectory-level rewards, while format reward measures compliance with the required output format. For skill compliance, skill-selection accuracy measures whether the required skill is correctly selected for each query, and the skill-adherence score evaluates how faithfully the resulting trajectory follows the prescribed skill execution procedure.

At the SFT stage, Reward + Entropy already leads on task performance, attaining the highest raw score and format reward together with the joint-highest skill-selection accuracy. The benefit of routing then compounds through RL: after RL, task performance and skill-adherence reward follow a consistent ordering, with Reward + Entropy outperforming Reward + Random, which in turn outperforms Random. Both routed arms improve consistently from SFT to RL, whereas Random shows no consistent improvement, indicating that reward routing provides learning signals that RL can exploit and that entropy prioritization contributes additional value within routed pools. Skill-selection accuracy differs only slightly across the three arms; the principal skill-compliance gain therefore lies in procedural adherence rather than skill selection.

\subsection{Reward Evolution Evaluation}

The results in Table~2 and Appendix Tables~2 and~3 show that the advantage of G-CAST becomes most visible once the capability contract expands to require explicit skill-dependent behavior, making later iterations the clearest setting for evaluating the reward-evolution loop. More generally, however, reward evolution in G-CAST is not only about tracking contract expansion; it keeps the executable reward reliable as a discriminator between desirable and undesirable trajectories under the current deployment setting.

We evaluate reward evolution from two complementary angles. First, we examine whether reward reliability problems arise in practice and therefore require governed reward updates rather than a fixed reward. Second, we test whether the benefit comes from governed reward evolution itself rather than from merely appending extra reward terms after the contract expands.

We begin by testing whether reward evolution is necessary once reward reliability starts to drift. In production, reward errors can appear not only as low-score failures but also as high-score trajectories that look plausible while violating important requirements. We therefore audit sampled high-reward trajectories instead of assuming that high reward implies correctness. This audit reveals two recurring high-score failure clusters, with false-positive rates of 84\% for search-result misuse and 82\% for hallucinated answer entities. In both cases, the agent invokes a plausible retrieval tool but still produces offer identifiers, URLs, prices, or phone numbers unsupported by the returned evidence. This finding motivates a penalty-only grounding component that checks whether verifiable entities in the final answer are supported by the prompt or tool responses. Before entering training, the new component must pass target-case tests, regression checks on known-good trajectories, frozen-trajectory rescoring, and replay. Reward evolution in G-CAST is therefore triggered by observed reward reliability failures and controlled by executable gates, rather than by unchecked reward expansion.

\begin{table}[h]
\centering
\small
\setlength{\tabcolsep}{3.5pt}
\begin{tabular}{@{}lccc@{}}
\toprule
Metric & Static & Ungated & Gated \\
\midrule
Cumulative reward            & 0.8810 & 0.8926 & \textbf{0.9288} \\
Period 1 reward           & 0.9248 & \textbf{0.9380} & 0.9336 \\
Period 2 reward           & 0.8381 & 0.8486 & \textbf{0.8990} \\
Period 3 reward           & 0.8801 & 0.8812 & \textbf{0.9538} \\
Format reward             & 0.9055 & 0.9025 & \textbf{0.9143} \\
Skill-selection accuracy  & 0.6867 & 0.6956 & \textbf{0.8667} \\
Skill adherence           & 0.4508 & 0.4865 & \textbf{0.5233} \\
\bottomrule
\end{tabular}
\caption{Comparison of training-reward designs on the cumulative evaluation over Periods 1–3.
All three arms share the same training data and optimization settings, and differ only in reward evolution under the expanded capability contract: Static keeps the original reward unchanged, Ungated introduces a one-time contract-aware reward update without continued gate-controlled evolution, and Gated (ours) admits reward updates only through continued validation gates.}
\label{tab:reward-design-ablation}
\end{table}

Second, we test whether the benefit comes from governed reward evolution rather than from merely adding extra reward terms. Table 4 reports a three-arm RL diagnostic from one shared SFT checkpoint. All three arms use the same training data and non-reward hyper-parameters, and differ only in how the training reward is handled after the capability contract expands to require explicit skill-dependent behavior. Static performs no reward evolution: it keeps the pre-expansion reward unchanged, so the newly introduced skill requirements are not reflected in training. Ungated applies a one-time contract-aware reward update, but does not continue evolving or re-admitting the reward through validation gates afterward. Gated (ours) continues to evolve the reward with the contract, and each reward update is allowed to affect training only after passing the registered target-case, regression, rescoring, and replay gates.

The comparison shows that simply adding more reward terms is not enough. Static reward establishes the lower bound when the training objective does not evolve with the contract, reaching 0.8810 overall reward and 0.8801 on the skill-intensive Period~3 evaluation. Ungated contract reward improves some aggregate metrics, reaching 0.8926 overall reward, but does not reliably improve the target behavior itself: skill-selection accuracy rises only slightly to 0.6956 and skill adherence to 0.4865. In contrast, gated contract-evolved reward reaches the best overall reward (0.9288), the best Period~3 reward (0.9538), the best skill-selection accuracy (0.8667), and the best skill-adherence score (0.5233). Relative to the static reward, the gain is especially pronounced where the contract expanded most: +0.0478 overall and +0.0737 on Period~3.

This result explains the crossover observed in the main continual-adaptation study. AutoTrainess is effective at broad post-training under a fixed objective, which is consistent with its competitive performance in Iterations~1--2. However, once the harness begins to require explicit Skill invocation and procedural compliance, training quality depends on whether the reward reflects the updated contract. A reward that does not evolve with the contract can still improve new-period reward to some extent, but it provides weaker supervision for the newly required behavior. G-CAST instead updates reward together with the contract, and the gains appear exactly where the contract changes matter most: skill selection, skill adherence, and retention under continued adaptation.

Taken together, these studies show that reward evolution is not an auxiliary implementation detail. It is a necessary part of continual adaptation under evolving demands. The benefit of G-CAST in later iterations comes not from using a larger reward, but from using a governed reward that remains valid for the capabilities the deployed agent is now expected to use.

\section{Conclusion and Discussion}

\paragraph{Operational Impact.}
During deployment in Yuancai, G-CAST reduced human reward-audit effort from about 2 person-day to 30 minutes per round and shortened demand-to-acceptance turnaround from about one week to two days. Its promotion and reward gates also rejected 3 candidate checkpoints and 2 proposed reward versions that failed validation, with rejection cases documented in Appendix E. These indicators quantify the value of governance itself, separately from the gains of any individual training run.

\paragraph{Limitations.}

Our study is based on one longitudinal production application and one primary
model family, so the observed patterns of demand and capability evolution may
not transfer directly to other agent settings.

\paragraph{Conclusion.}
As deployed e-commerce agents face changing user demands, their capability contracts also evolve through new tools, skills, and operating procedures, making continual adaptation a coupled problem of model mismatch, data mismatch, and reward mismatch rather than repeated training under a fixed objective. To address this, we introduced G-CAST, a governed continual self-training system that treats the checkpoint transition as the unit of adaptation and jointly manages model updates, reward validity, routed learning evidence, and cumulative evaluation. By combining reward-routed, entropy-prioritized data orchestration with capability-contract-driven reward evolution, G-CAST enables the agent to learn newly introduced behaviors while preserving previously accepted capabilities. Experiments on a real production e-commerce agent show that this formulation is both necessary and effective: a fixed checkpoint degrades substantially as the contract expands, whereas G-CAST achieves stronger adaptation, better retention, and more reliable improvement under cumulative deployment constraints.

\clearpage
\bibliography{references}

\clearpage

\end{document}